\documentclass[letterpaper, 10 pt, conference]{ieeeconf}  % Comment this line out if you need a4paper

\IEEEoverridecommandlockouts                              % This command is only needed if
\title{\LARGE \bf
DROW: Real-Time Deep Learning based Wheelchair Detection in 2D Range Data
}

\author{Lucas Beyer$^{\dagger}$, Alexander Hermans$^{\dagger}$ and Bastian Leibe% <-this % stops a space
\thanks{\hspace*{-3.5pt}$^{\dagger}$Equal contribution. Ordering determined by a last minute coin flip.}
\thanks{The work in this paper was funded by the EU project STRANDS (ICT-2011-600623).
        All authors are at the Visual Computing Institute, RWTH Aachen University.
        e-mail: {\tt\footnotesize last@vision.rwth-aachen.de}}%
}

\usepackage{makeidx}
\usepackage{graphicx}
\usepackage{subcaption}
\usepackage{wrapfig}
\usepackage{amsmath,amssymb} % define this before the line numbering.
\usepackage{color}
\usepackage{listings}
\usepackage{hyperref}
\usepackage{xspace}
\usepackage{tabularx}
\usepackage{tikz}
\usetikzlibrary{calc,shapes,arrows}
\usepackage{pgfplots}
\usepackage{siunitx}
\usepackage[bottom]{footmisc}

\pgfplotsset{precrecbase/.style={
              inner sep=0pt,outer sep=0pt,
              ylabel style={font=\scriptsize,yshift=-18pt},
              xlabel style={font=\scriptsize,yshift=6pt},
              width={1.1\textwidth},
              height={5cm},
              yticklabel style = {font=\scriptsize,xshift=-0.3ex},
              xticklabel style = {font=\scriptsize,yshift=-0.3ex},
              legend image post style={line width =1.5pt}
              }
}
\pgfplotsset{precrec/.style={
              precrecbase,
              xtick={0,20,...,100},
              ytick={0,20,...,100},
              xlabel={Recall [\%]},
              ylabel={Precision [\%]},
              xmin=-2,
              xmax=102,
              ymin=-2,
              ymax=102,
              grid=both,
              grid style={dashed,gray!25!white}
              }
}

\pgfplotsset{legend image code/.code={%
                \draw[mark repeat=2,mark phase=2]
                plot coordinates {
                (0cm,0cm)
                (0.3cm,0cm)        %% default is (0.3cm,0cm)
                (0.6cm,0cm)         %% default is (0.6cm,0cm)
                };%
            }
}

\newcommand{\rf}[1]{{\protect\NoHyper\ref{#1}\protect\endNoHyper}}

\definecolor{wc_c}{HTML}{348ABD}
\definecolor{wa_c}{HTML}{D175F0}
\definecolor{laser_red}{HTML}{E24A33}
\definecolor{bg_green}{HTML}{8EBA42}
\definecolor{awesome_orange}{HTML}{F9A91F}
\definecolor{good_gray}{HTML}{777777}
\definecolor{deep_pink}{HTML}{FF69B4}
\definecolor{shit_brown}{HTML}{8B4513}
\definecolor{random_blue}{HTML}{7DE2F0}

\makeatletter
\DeclareRobustCommand\onedot{\futurelet\@let@token\@onedot}
\def\@onedot{\ifx\@let@token.\else.\null\fi\xspace}
 
\def\ie{\emph{i.e}\onedot} 
\def\cf{\emph{c.f}\onedot} 
\def\etc{etc\onedot} 
 
\def\etal{\emph{et al}\onedot}
\makeatother

\newcommand{\mytt}[1]{{\tt \footnotesize #1}}

\newcommand{\pb}[3]{%
    \hspace*{-7pt}
    \begin{tikzpicture}[baseline={($ (current bounding box.north) - (0,8pt) $)}]
        \begin{axis}[axis lines=none,
                     ticks=none,
                     xticklabel=\empty,
                     yticklabel=\empty,
                     height=2.3cm,
                     width=3.15cm,
                     ybar, ybar interval]
            \addplot [color=#1, fill=#1, fill opacity=0.4] table [x=dist, y=#2] {data/#3.dat};
        \end{axis}
    \end{tikzpicture}%
        \vspace*{1pt}
}

\newcommand{\lc}[1]{\multicolumn{1}{c}{#1}}

\newenvironment{customlegend}[1][]{%
    \begingroup
    \csname pgfplots@init@cleared@structures\endcsname
    \pgfplotsset{#1}%
}{%
    \csname pgfplots@createlegend\endcsname
    \endgroup
}%
\def\addlegendimage{\csname pgfplots@addlegendimage\endcsname}

\begin{document}

    \maketitle
    \thispagestyle{empty}
    \pagestyle{empty}

    %%%%%%%%%%%%%%%%%%%%%%%%%%%%%%%%%%%%%%%%%%%%%%%%%%%%%%%%%%%%%%%%%%%%%%%%%%%%%%%%
    \begin{abstract}
    We introduce the DROW detector, a deep learning based object detector operating on 2D range data.
    Laser scanners are lighting invariant, provide accurate 2D range data, and typically cover a large field of view, making them interesting sensors for robotics applications.
    So far, research on detection in laser 2D range data has been dominated by hand-crafted features and boosted classifiers, potentially losing performance due to suboptimal design choices.
    We propose a Convolutional Neural Network (CNN) based detector for this task.
    We show how to effectively apply CNNs for detection in 2D range data, and propose a depth preprocessing step and a voting scheme that significantly improve CNN performance.
    We demonstrate our approach on wheelchairs and walkers, obtaining state of the art detection results.
    Apart from the training data, none of our design choices limits the detector to these two classes, though.
    We provide a ROS node for our detector and release our dataset containing 464k laser scans, out of which 24k were annotated.
    \end{abstract}

    %%%%%%%%%%%%%%%%%%%%%%%%%%%%%%%%%%%%%%%%%%%%%%%%%%%%%%%%%%%%%%%%%%%%%%%%%%%%%%%%
    \section{INTRODUCTION}
    Many autonomous robots are equipped with a 2D laser scanner, typically used for navigation-related tasks including the detection of people~\cite{Arras07ICRA,Weinrich14ROMAN} and objects~\cite{Merdrignac15ARSO}.
    Laser scanners are widely used due to their typically large field of view and their invariance to lighting and environmental conditions.
    While early detection methods used simple heuristics such as fitting lines and circles~\cite{Xavier05ICRA}, in the past few years hand crafted features, coupled with learned classifiers, have dominated laser based detection.
    Within this paradigm, a range of successful people~\cite{Arras07ICRA,Spinello08ICRA,Mozos10IJSR}, mobility aid~\cite{Weinrich14ROMAN} and road obstacle~\cite{Merdrignac15ARSO} detectors have been developed to support mobile robot navigation~\cite{Dondrup15ICRA} and autonomous driving~\cite{Sivaraman13TITS}.
    Even though the aforementioned models obtain respectable results, the general consensus seems to be that the information provided by 2D range data is not sufficient to reliably perform detection in a single scan, leading to approaches relying on sensor fusion~\cite{Spinello08ICRA}, multi-layered sensor setups~\cite{Mozos10IJSR,Spinello10AAAI}, or temporal integration of information by tracking.
    We show that detection based on single 2D range scan can actually perform well.

    In this paper, we specifically focus on the detection of wheelchairs and walkers.
    This is motivated by a service robot application in an elderly care facility, where many people rely on their mobility aids.
    Since the presence of a mobility aid can significantly change a person's appearance, both in laser and in camera data, they are less reliably detected by existing people detectors. 
    However, especially people relying on those aids will have a harder time avoiding an approaching robot, making a reliable detection all the more important.
    %thus feel intimidated if they cannot ensure a safe distance between them and a robot
    Weinrich~\etal~\cite{Weinrich14ROMAN} face a similar task and propose the Gandalf detector for detecting people, wheelchairs, and walkers in individual laser scans.
    They introduce a new feature set for laser segments and classify these with an AdaBoost classifier.
    The resulting detector performs reasonably well and we include it as a baseline.

    \begin{figure}
    \tikzstyle{laser} = [draw, circle, fill=yellow,scale=5,line width=2pt]
    \tikzstyle{point} = [circle, fill=laser_red, fill opacity=0.88,scale=2.5]
    \tikzstyle{ann} = [inner sep=1pt,scale=5]
    \definecolor{window_c}{HTML}{38942A} %{8AE27C}
    \centering
    \resizebox {\columnwidth} {!} {
    \begin{tikzpicture}
    \node [laser] (laser) {};

    \pgfplotstableread{data/scan.dat}{\scanpoints}
    \pgfplotstablegetrowsof{data/scan.dat}
    \pgfmathsetmacro{\rows}{\pgfplotsretval-2}
    \foreach \j in {0,...,\rows}{%
        \pgfplotstablegetelem{\j}{[index] 0}\of{\scanpoints}
        \let\x\pgfplotsretval
        \pgfplotstablegetelem{\j}{[index] 1}\of{\scanpoints}
        \let\y\pgfplotsretval
        \node [point] (p) at  (\x,\y) {};
        \draw [color=laser_red, opacity=0.25, line width=4pt , -] (laser) -- (p);
    }%
    \pgfplotstablegetrowsof{data/scan.dat}
    \pgfmathsetmacro{\last}{\pgfplotsretval-1}
    \pgfplotstablegetelem{\last}{[index] 0}\of{\scanpoints}
    \let\xd\pgfplotsretval
    \pgfplotstablegetelem{\last}{[index] 1}\of{\scanpoints}
    \let\yd\pgfplotsretval

    \draw [-angle 60,line width=3pt] (0,0) -- (0,3) node[right, ann] {$y^w$};
    \draw [-angle 60,line width=3pt] (0,0) -- (3,0) node[below, ann] {$x^w$};

    %Wc detection
    \node [text=wc_c, scale=9] (det)  at  (\xd,\yd) {+};
    \node[draw, circle,color=wc_c, fill=wc_c, scale=40, draw opacity=0.25, fill opacity=0.1,line width=2pt]  at  (\xd,\yd) {};

    %Angle at the bottom
    \draw [line width=3pt, angle 60-angle 60] (-3.46370263,3.60593456) arc (133.847:101.776:5);
    \node [ann] at (-3.17,4.77) {$\alpha$};

    %Helper lines in black
    \draw [-] (laser) -- (-3.87768927,18.60009576);
    \draw [-] (laser) -- (-13.1620707,13.702552);
    \draw [-,line width=1pt] (laser) -- (-13.079775534591909,25.88280369816613);
    \draw [line width=2pt, -] (-20.0894759,20.9144211) arc (133.847:101.776:29);

    %Green box
    \draw [line width=4pt, -, color=window_c] (-3.87768927,18.60009576) -- (-7.9594673,38.1791429);
    \draw [line width=4pt, -, color=window_c] (-13.1620707,13.702552) -- (-27.0168812,28.1262903);
    \draw [line width=4pt, color=window_c] (-27.0168812,28.1262903) arc (133.847:101.776:39);
    \draw [line width=4pt, color=window_c] (-13.1620707,13.702552) arc (133.847:101.776:19);

    %Local offset
    \draw [-angle 60,line width=3pt, color=window_c] (-13.079775534591909,25.88280369816613) -- (-14.133857034860622,27.968664000674353) {};
    \draw [-angle 60,line width=3pt, color=window_c] (-13.079775534591909,25.88280369816613) -- ++(3.3111318660327921,1.6732677834361476) {};
    \draw [loosely dashed,line width=2pt, color=window_c] (-14.133857034860622,27.968664000674353) --  (\xd,\yd) node[midway, ann, shift={(-0.1,0.25)}, color=window_c]  {$\Delta x$} ;
    \draw [loosely dashed,line width=2pt, color=window_c] (-9.7686436685591169,27.5560714816022776) --  (\xd,\yd) node[right,midway, ann, shift={(0.0,0.05)},color=window_c] {$\Delta y$};

    %Black measures at the side.
    \draw [line width=3pt, angle 60-angle 60] (-13.6620707,13.202552) --  (-20.62447594,20.44842114) node[below, left, midway, ann, shift={(-0.1,-0.1)}] {$H_r$};
    \draw [line width=3pt, angle 60-angle 60] (-20.55947594,20.38442114) -- (-27.5168812,27.6262903) node[below, left, midway, ann, shift={(0.0,-0.29)}] {$H_r$};

    \node [fill=white,scale=5,draw,rectangle, rounded corners=0.4cm,line width = 4pt, inner sep = 0.2cm] at (-39, 2) {CNN};
    \node [fill=white,scale=5,draw,rectangle, rounded corners=0.4cm,line width = 4pt, inner sep = 0.2cm] at (-39.5, 1.5) {CNN};
    \node [fill=white,scale=5,draw,rectangle, rounded corners=0.4cm,line width = 4pt, inner sep = 0.2cm] at (-40, 1) {CNN};

    \draw [loosely dashed,line width=4pt, -angle 60] (-39.5, 6.5) -- (-39.5, 4) {};
    \draw [loosely dashed,line width=4pt, -angle 60] (-39, 7) -- (-39, 4.5) {};
    \draw [loosely dashed,line width=4pt, -angle 60] (-40, 6) -- (-40, 3.5) {};

     \begin{axis}[ axis background/.style={fill=white},
                   scale=2,
                   xtick style={draw=none},
                   ytick style={draw=none},
                   xticklabels={,,},
                   yticklabels={,,},
                   xmin=-1,
                   xmax=48,
                   ymin=-1,
                   ymax=1,
                   line width = 4pt,
                   window_c,
                   at={(-46cm, 7.5cm)}]
             \addplot [color=black, mark=none, line width = 2pt] coordinates {(-1,0) (48,0)};
     \end{axis}
     \begin{axis}[ axis background/.style={fill=white},
                   scale=2,
                   xtick style={draw=none},
                   ytick style={draw=none},
                   xticklabels={,,},
                   yticklabels={,,},
                   xmin=-1,
                   xmax=48,
                   ymin=-1,
                   ymax=1,
                   line width = 4pt,
                   window_c,
                   at={(-46.5cm, 7cm)}]
            \addplot [color=black, mark=none, line width = 2pt] coordinates {(-1,0) (48,0)};
    \end{axis}
    \begin{axis}[ axis background/.style={fill=white},
                  scale=2,
                  xtick style={draw=none},
                  ytick style={draw=none},
                  xticklabels={,,},
                  ytick={-1,0,1},
                  yticklabels={\resizebox{!}{1.2cm}{$H_r$}, \resizebox{!}{1.2cm}{$0$}, \resizebox{!}{1.2cm}{$\text{-}H_r$}},
                  yticklabel style = {black},
                  xmin=-1,
                  xmax=48,
                  ymin=-1,
                  ymax=1,
                  line width = 4pt,
                  window_c,
                  at={(-47cm, 6.5cm)}]
             \addplot [color=black, mark=none, line width = 2pt] coordinates {(-1,0) (48,0)};
             \addplot [ mark options={scale=2, fill=laser_red}, color=laser_red, mark=*, only marks] table [y=y, x=x, col sep=comma] {data/window.csv};
           \end{axis}

    \node [scale=5,draw,rectangle, rounded corners=0.4cm,line width = 4pt, inner sep = 0.17cm] at (-31, 1.5) {Voting};
    \node [scale=5,draw,rectangle, rounded corners=0.4cm,line width = 4pt, inner sep = 0.2cm] at (-22.5, 1.5) {NMS};
    \node [wc_c, fill = wc_c, fill opacity=0.25, text=wc_c, text opacity=1,     scale=5,draw,rectangle, rounded corners=0.4cm,line width = 4pt, inner sep = 0.2cm] at (-12.2, 1.5) {Detections};

    \draw [loosely dashed,line width=4pt, angle 60-] (-30.5, 13) arc (280:325:14cm) {};
    \draw [loosely dashed,line width=4pt] (-36, 2) -- (-34.5, 1.5) {};
    \draw [loosely dashed,line width=4pt] (-37, 1) -- (-34.5, 1.5) {};
    \draw [loosely dashed,line width=4pt, -angle 90] (-36.5, 1.5) -- (-34.5, 1.5) {};
    \draw [loosely dashed,line width=4pt, -angle 60] (-27.5, 1.5) -- (-25.8, 1.5) {};
    \draw [loosely dashed,line width=4pt, -angle 60] (-19.2, 1.5) -- (-17.5, 1.5) {};
    \end{tikzpicture}}
    \caption{An overview of our approach. We extract windows with a fixed real world size regardless of the laser coverage.
    Values are centered and clamped to the range of the window.
    Each window is classified by a CNN and, if an object is found, a weighted vote is cast for its relative offset $(\Delta x, \Delta y)$.
    Finally, the votes are consolidated using a non-maximum suppression (NMS) scheme to yield detection centroids.}
    \label{fig:depth_prepocessing}
    \end{figure}
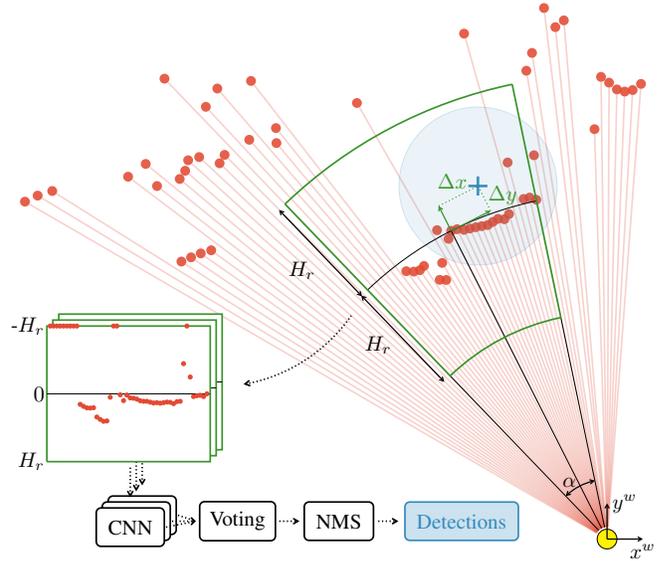

    \newcommand{\rulesep}{\unskip\ \textcolor{gray!50}{\vrule width 1.2pt}\,}
    %Needs to be here because fuck you latex!
    \begin{figure*}[!ht]
    \centering
    \begin{subfigure}{0.155\textwidth}
        \includegraphics[width=\textwidth]{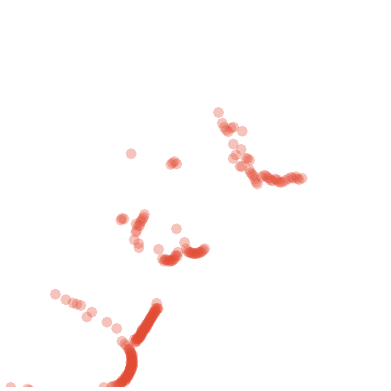}
    \end{subfigure}%
    \rulesep
    \begin{subfigure}{0.155\textwidth}
        \includegraphics[width=\textwidth]{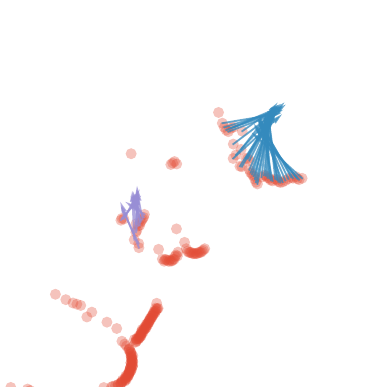}
    \end{subfigure}%
    \rulesep
    \begin{subfigure}{0.155\textwidth}
        \includegraphics[width=\textwidth]{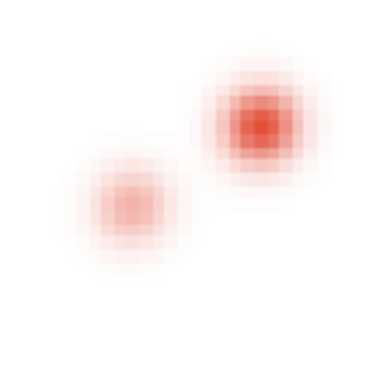}
    \end{subfigure}%
    \rulesep
    \begin{subfigure}{0.155\textwidth}
        \includegraphics[width=\textwidth]{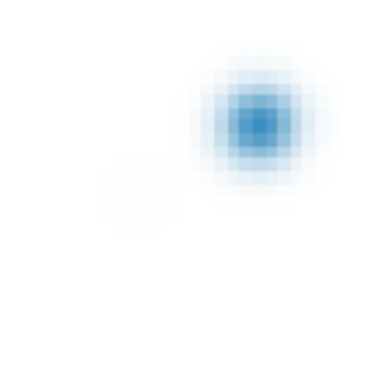}
    \end{subfigure}%
    \rulesep
    \begin{subfigure}{0.155\textwidth}
        \includegraphics[width=\textwidth]{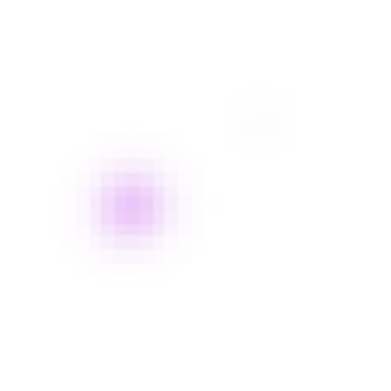}
    \end{subfigure}%
    \rulesep
    \begin{subfigure}{0.155\textwidth}
        \includegraphics[width=\textwidth]{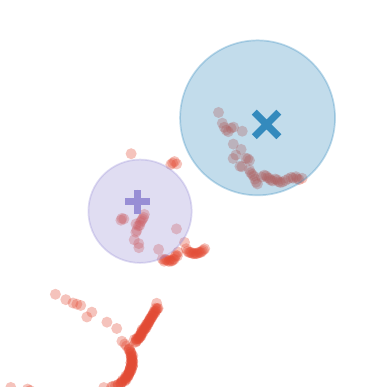}
    \end{subfigure}%

    \vspace*{-10pt}
    \begin{subfigure}{0.1663\textwidth}
    \makebox[\textwidth][r]{(a)\hspace{4pt}}
    \end{subfigure}%
    \begin{subfigure}{0.1663\textwidth}
    \makebox[\textwidth][r]{(b)\hspace{4pt}}
    \end{subfigure}%
    \begin{subfigure}{0.1663\textwidth}
    \makebox[\textwidth][r]{(c)\hspace{4pt}}
    \end{subfigure}%
    \begin{subfigure}{0.1663\textwidth}
    \makebox[\textwidth][r]{(d)\hspace{4pt}}
    \end{subfigure}%
    \begin{subfigure}{0.1663\textwidth}
    \makebox[\textwidth][r]{(e)\hspace{4pt}}
    \end{subfigure}%
    \begin{subfigure}{0.1663\textwidth}
    \makebox[\textwidth][r]{(f)\hspace{4pt}}
    \end{subfigure}%
    \caption{Non-maximum suppression: (a) a part of an input scan, (b) the votes shown as arrows, (c) the joint voting grid, (d) the wheelchair voting grid, (e) the walker voting grid, (f) resulting detections.}
    \label{fig:nms}
    \end{figure*}

    In this paper, we introduce the DROW (Distance RObust Wheelchair/Walker) detector.
    Driven by our application scenario, we focus on the detection of wheelchairs and walkers.
    However, there are no design decisions which restrict our approach to these objects and we are confident it will generalize to detection of persons or other objects, given sufficient annotated training data.
    Code and data needed to train and run our detector, as well as code to annotate data for new tasks, will be made available upon publication.

    In computer vision, deep learning has recently become the new best practice, replacing hand-crafted features by learned ones and overhauling the state of the art in many tasks~\cite{Krizhevsky12NIPS,Tompson14NIPS,Beyer15GCPR}.
    Specifically, Convolutional Neural Networks (CNNs) have been very successful at challenging tasks.
    In this paper, we show how CNNs can be applied for object detection in laser data, alleviating the need for feature engineering and enabling drastic improvements.
    
    While CNN-based image-level detectors like MultiBox~\cite{Liu15Arxiv} and YOLO~\cite{Redmon15Arxiv} could in principle be applied to a 2D laser scan, we found that doing so naively is not effective.
    Since the spatial density of laser points varies greatly with distance, the fixed perceptive field of a CNNs covers widely different scales, making learning difficult.
    To make use of the spatial information a laser sensor provides, we propose a preprocessing stage that cuts out and normalizes a fixed real-world extent window around each laser point.
    All of those windows are fed through a CNN which can cast votes for object locations.
    These votes are then turned into individual detections using a non-maximum suppression scheme.
    We show that both depth preprocessing and voting are essential components of our approach, an overview of which can be seen in Fig.\,\ref{fig:depth_prepocessing}.
    Our approach does not require background subtraction and runs with a frame rate of $\sim$75\,fps on a modern desktop machine using a single core and a GPU.
    On our robot, it easily keeps up with the laser frame rate ($\sim$13\,fps) on a laptop GPU.

    To summarize, we make the following contributions:
    \begin{itemize}
        \item We introduce the DROW detector, a CNN based wheelchair and walker detector for 2D range data which, by effectively making use of the provided depth-information, achieves state of the art results.
        \item We publish our dataset, containing 464k raw scans of which 24k have been annotated with wheelchair and walker centroids.
        \item We provide ROS components of our detector and related service modules, including trained models.
    \end{itemize}

    %%%%%%%%%%%%%%%%%%%%%%%%%%%%%%%%%%%%%%%%%%%%%%%%%%%%%%%%%%%%%%%%%%%%%%%%%%%%%%%%
    \section{APPROACH}
    \label{sec:approach}

    Our approach consists of three steps: preprocessing, which cuts out a resampled window around every laser point and computes detection locations in a local coordinate system, a CNN classifying said windows and predicting relative detection locations, and finally a voting and non-maximum suppression scheme turning predictions into detections.

    \subsection{2D Range Data Preprocessing}\label{sec:approach_preproc}
    In order to apply a CNN for detection, the network's receptive field must cover a large part of the object.
    The problem with laser scans is that nearby objects cover a large amount of laser beams, whereas distant objects are only hit by a handful of beams.
    This means the CNN's receptive field must cover most of the laser, which makes it very prone to overfitting to the training scenes' backgrounds.

    To circumvent this problem, and at the same time make use of the real-world scale information that laser data provides, we propose to evaluate the CNN in a depth-guided sliding-window fashion.
    This means that we preprocess the data such that objects have approximately the same representation at every distance, thus alleviating the need to implicitly learn completely different representations at varying distances:
    Around each beam, we cut out a window of real-world extent $\ell$, thus spanning an angle $\alpha=2\sin^{-1}({\ell\over2r})$ and containing a variable amount of measurements depending on the distance $r$ at which the current beam hits an obstacle.
    We then resample the measurements inside this window linearly at $n$ fixed samples.
    When applied to such a window, the network's receptive field always covers the same real-world extent, regardless of the distance $r$.
    In addition, we center the window around the current point, clamp any values outside a $\pm H_r$ hull in order to remove distant clutter and finally project the values into $[-1,1]$.
    In total, this means $\max(-H_r, \min(x-r, H_r))/H_r$ is applied to each point $x$ of the window around the point at depth $r$, as illustrated in Fig.\,\ref{fig:depth_prepocessing}.
    A detailed analysis regarding which of these preprocessing operations contribute the most to DROW's performance can be found in Section~\ref{sec:ablation}.

    \subsection{Prediction}
    For each window, and thus for each laser point with its context, a CNN both classifies whether this window belongs to an object class of interest through a SoftMax output and, if so, votes for the center location of that object through a regression output.
    As the 2D range data is inherently rotation-invariant, we do not want to perform voting in absolute $(x,y)$ coordinates.
    Instead, we learn offsets $(\Delta x, \Delta y)$ in a coordinate system centered and aligned at the current laser point, as shown in Fig.\,\ref{fig:depth_prepocessing}.

    \subsection{Voting and Non-Maximum Supression}\label{sec:vote_nms}
    The predictions for every window need to be consolidated into detection centers.
    This is achieved by making the CNN's predictions vote into regular grids spanning the laser's field of view.
    Let $p(O|w) = \sum_{c \in \mathcal{C}} p(c|w)$ be the total probability of the window $w$ seeing an object of interest, where $\mathcal{C}$ are the classes to be detected.

    If $p(O|w)$ exceeds a predefined voting threshold $T$, the window will cast a vote into a class-agnostic grid with weight $p(O|w)$ as well as into each class-specific grid with weight $p(c|w)$.
    After all windows have potentially cast votes, each grid is blurred with a Gaussian filter and non-maximum suppression is performed on the class-agnostic grid.
    For each maximum found in that grid, a detection is predicted at the cell's center using the class which has the highest sum of votes in said cell.
    The reason for this voting scheme, as opposed to treating each class separately, is to avoid detections of both classes at the same position.
    Fig.\,\ref{fig:nms} shows an example of the different steps involved. Votes cast from the raw laser points in (a) are shown in (b); (c) - (e) show the three voting grids and (f) shows the two resulting detections.

    %%%%%%%%%%%%%%%%%%%%%%%%%%%%%%%%%%%%%%%%%%%%%%%%%%%%%%%%%%%%%%%%%%%%%%%%%%%%%%%%
    \section{EXPERIMENTAL EVALUATION}
    For the evaluation of our approach we first introduce our new dataset.
    We outline the details of our training procedure and of the evaluation methods.
    After the general evaluation of DROW and a comparison to baselines, we perform several ablation studies to show how much individual parts of our detector contribute to the overall performance.

    \subsection{Dataset}
    Although Weinrich~\etal~\cite{Weinrich14ROMAN} provide their datasets, their recordings are limited to scenes with a single wheelchair or walker.
    While this might suffice for learning the few parameters of fairly constrained features and classifiers, we want to learn features from scratch and thus require more general and varied data.
    This is why we recorded a little over 10 hours of data at an elderly care facility.

    \subsubsection{Recording Setup}
    \begin{wrapfigure}[17]{l}{0.4\linewidth}
      \vspace*{-10pt}
      \begin{tikzpicture}
        \node at (0,2.5) {\includegraphics[width=\linewidth]{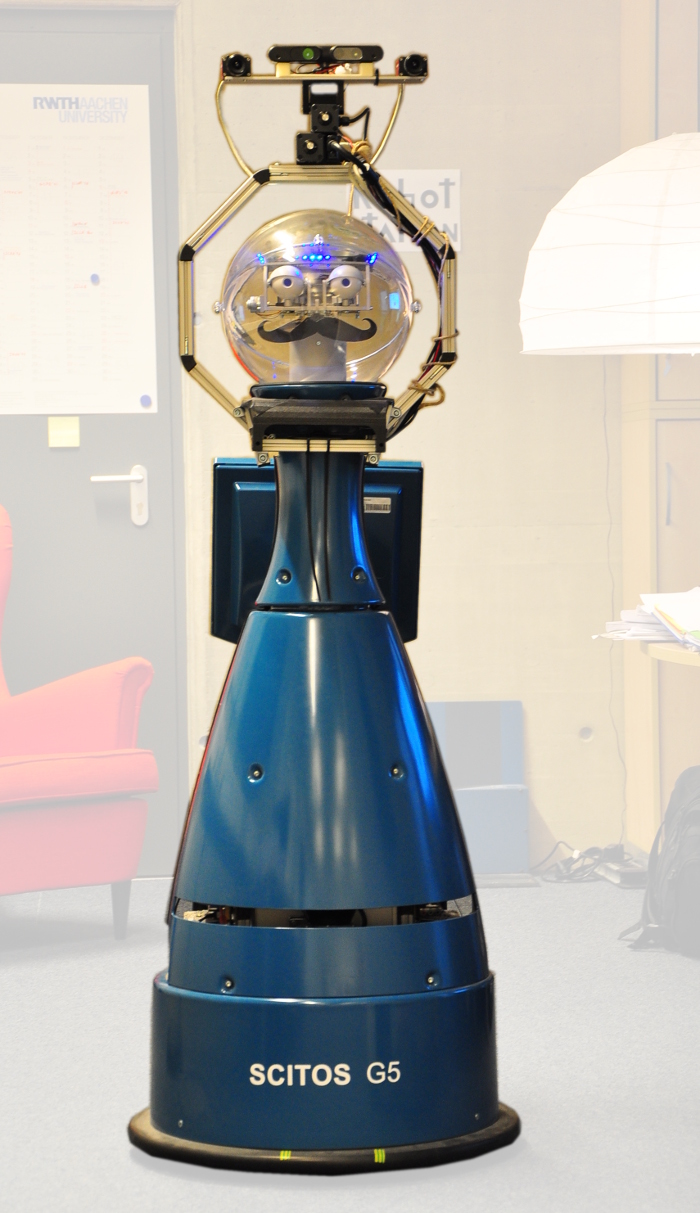}};
        \draw [angle 60-angle 60] (1,-0.15) -- (1,5.2) node[midway, right,anchor=north, rotate=90] {\scriptsize \SI{170}{\cm}};
        \draw [angle 60-angle 60] (1.2,-0.15) -- (1.2,0.97) node[midway, right,anchor=north, rotate=90] {\scriptsize \SI{37}{\cm}};
      \end{tikzpicture}
      \caption{Our robot Karl.}
      \label{fig:karl}
    \end{wrapfigure}
    We recorded the data using a SCITOS G5 robot equipped with a SICK S300 laser scanner mounted at approximately \SI{37}{\cm} above the ground plane.
    The laser was configured to record at nearly \SI{13}{\Hz} and span a field of view of \SI{225}{\degree} at a resolution of \SI{0.5}{\degree}, totalling $450$ measurements.
    For annotation purposes, we also recorded RGB-D data from an ASUS Xtion mounted on the robot's head~(see Fig.\,\ref{fig:karl}).
    Unfortunately, we are not allowed to publish the video streams for privacy reasons.
    The software infrastructure is based on ROS and all data was stored in rosbags.
    To ensure seeing enough wheelchairs and walkers, one person permanently roamed around using either while the robot was recording.
    Our recordings consist of both (1) natural scenes where we recorded the everyday life in the facility including all kinds of clutter such as flower pots, chairs, furniture, rolling beds,~\ldots,
    and (2) artificial recordings where we drove around in certain patterns as to bring variation to the dataset.
    Apart from the resident's wheelchairs, we included as many wheelchair and walker models as possible, including one motorized wheelchair.

    \subsubsection{Dataset Statistics}
    We split the resulting dataset into a train, validation and test set.
    To create these sets, we split the care facility into four non-overlapping areas, three of which were assigned to the train, test and validation sets, respectively.
    The fourth, the entrance hall, was split into temporally disjoint sequences which were distributed over the train and validation sets.
    Based on this split, we can show how well the approach generalizes to never before seen areas.
    Table~\ref{table:dataset} shows an overview of our dataset, as well as statistics of the subset we annotated.
    The wheelchair and walker counts refer to individual detections as opposed to instances, and the bar plots show their distribution over the distance.
    Each bar represents a \SI{1}{\m} slice in the distance (15 bars for up to \SI{15}{\m}), clearly showing that the majority of observed mobility aids are encountered within \SIrange{1}{6}{\m} of the robot.
    \newcolumntype{Y}{>{\raggedleft\arraybackslash}X}
        \begin{table}[t]
        \caption{Dataset overview}
        \label{table:dataset}
        \begin{tabularx}{\linewidth}{p{1.9cm}YYYY}
            \hline
                                 & \lc{Train}    & \lc{Validation} & \lc{Test}    & \lc{Total}   \\\hline
             Sequences           & \lc{78}       & \lc{30}         & \lc{5}       & \lc{113}     \\
             Scans               & 341\,138\,\,\, & 74\,744\,\,\,    & 48\,131\,\,\, & 464\,013\,\,\, \\
             Annotated Scans     & 17\,665\,\,\,  & 3\,919\,\,\,       & 2\,428\,\,\,    & 24\,012\,\,\,  \\
             Wheelchairs         & 14\,455\,\,\,  & 5\,595\,\,\,       & 1\,970\,\,\,    & 22\,020\,\,\,  \\
             Walkers             & 2\,047\,\,\,     & 219\,\,\,        & 581\,\,\,     & 2\,847\,\,\,   \\
             Wheelchairs\newline by distance &\pb{wc_c}{wc}{train}   & \pb{wc_c}{wc}{valid}  & \pb{wc_c}{wc}{test}   & \pb{wc_c}{wc}{all} \\
             Walkers\newline by distance & \pb{wa_c}{wa}{train}   & \pb{wa_c}{wa}{valid}   & \pb{wa_c}{wa}{test}    & \pb{wa_c}{wa}{all}  \\
            \hline
        \end{tabularx}
        \end{table}

    \subsubsection{Annotation}
    \begin{figure}[b]
        \centering%
        \includegraphics[width=\linewidth]{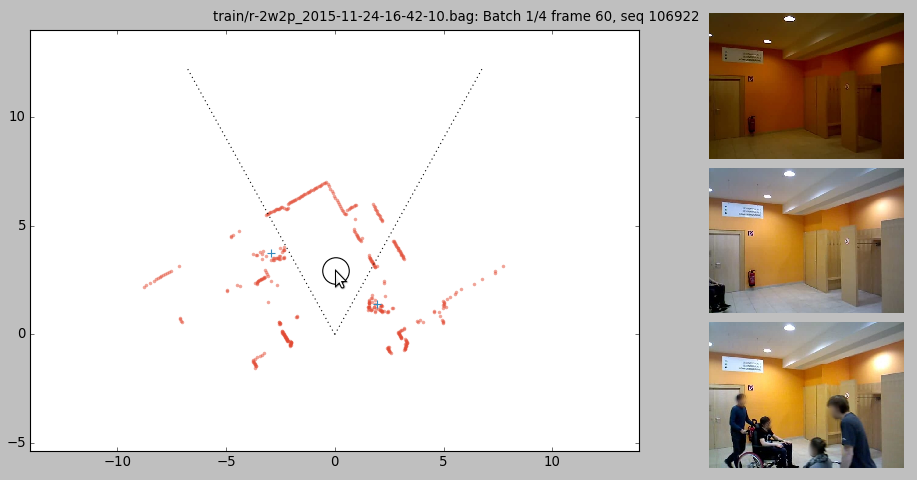}%
        \caption{An example view of our annotation tool. The dotted cone shows the Xtion's field of view and on the right the first, current, and last image in the batch is shown.
        The blue cross shows annotated wheelchairs.}%
        \label{fig:annotation_tools}%
    \end{figure}

    \begin{figure*}[t]
        \centering
        \begin{subfigure}{0.435\textwidth}
            \begin{tikzpicture}[]
            \begin{axis}[precrec, height=5.33cm]
            \addplot [thick, mark=none, color=laser_red] table [y=p_all_0.5, x=r_all_0.5, col sep=comma] {data/pr_test.csv};\label{all}
            \addplot [thick, mark=none, color=wc_c] table [y=p_wc_0.5, x=r_wc_0.5, col sep=comma] {data/pr_test.csv};\label{wc}
            \addplot [thick, mark=none, color=wa_c] table [y=p_wa_0.5, x=r_wa_0.5, col sep=comma] {data/pr_test.csv};\label{wa}
            \addplot [thick, dotted, mark=none, color=laser_red] table [y=p_all_0.5, x=r_all_0.5, col sep=comma] {data/pr_test_raw-det.csv};
            \addplot [thick, dotted, mark=none, color=wc_c] table [y=p_wc_0.5, x=r_wc_0.5, col sep=comma] {data/pr_test_raw-det.csv};
            \addplot [thick, dotted, mark=none, color=wa_c] table [y=p_wa_0.5, x=r_wa_0.5, col sep=comma] {data/pr_test_raw-det.csv};
            \addplot [mark=*, mark size=1pt, color=laser_red] table [y=p_all, x=r_all, col sep=comma] {data/pr_test_weinrich_0.5.csv};
            \addplot [mark=*, mark size=1pt, color=wc_c] table [y=p_wc, x=r_wc, col sep=comma] {data/pr_test_weinrich_0.5.csv};
            \addplot [mark=*, mark size=1pt, color=wa_c] table [y=p_wa, x=r_wa, col sep=comma] {data/pr_test_weinrich_0.5.csv};
            \addplot [mark=x, mark size=2pt, color=laser_red] table [y=p_all, x=r_all, col sep=comma] {data/pr_test_weinrich_0.5_p.csv};
            \end{axis}
            \end{tikzpicture}
            \caption{Performance on our test set.}
            \label{fig:drow_our_test}
        \end{subfigure}%
        \begin{subfigure}{0.435\textwidth}
            \begin{tikzpicture}
            \begin{axis}[precrec, height=5.33cm, ylabel={}, legend image post style={line width =.5pt}]
            \addplot [thick, mark=none, color=laser_red] table [y=p_all_0.5, x=r_all_0.5, col sep=comma] {data/pr_reha.csv};
            \addplot [thick, mark=none, color=wc_c] table [y=p_wc_0.5, x=r_wc_0.5, col sep=comma] {data/pr_reha.csv};
            \addplot [thick, mark=none, color=wa_c] table [y=p_wa_0.5, x=r_wa_0.5, col sep=comma] {data/pr_reha.csv};
            \addplot [mark=*, mark size=1pt, color=laser_red, only marks] table [y=p_all, x=r_all, col sep=comma] {data/pr_reha_weinrich_0.5.csv};\label{plot:gandalf}
            \addplot [mark=*, mark size=1pt, color=wc_c] table [y=p_wc, x=r_wc, col sep=comma] {data/pr_reha_weinrich_0.5.csv};
            \addplot [mark=*, mark size=1pt, color=wa_c] table [y=p_wa, x=r_wa, col sep=comma] {data/pr_reha_weinrich_0.5.csv};
            \addplot [mark=x, mark size=2pt, color=laser_red, only marks] table [y=p_all, x=r_all, col sep=comma] {data/pr_reha_weinrich_0.5_p.csv};\label{plot:gandalfp}
            \end{axis}
            \end{tikzpicture}
            \caption{Performance on the Reha test set~\cite{Weinrich14ROMAN}.}
        \end{subfigure}\hspace*{-5pt}%
        \begin{subfigure}{0.10\textwidth}
            \vspace*{-30pt}
            \begin{tikzpicture}
            \begin{customlegend}[every axis legend/.append style={nodes={right}}, legend style={draw=none,font=\scriptsize}, legend entries={Agnostic, Wheelchair, Walker,~,~,~,Drow ,Gandalf~\cite{Weinrich14ROMAN}, \parbox{2cm}{Gandalf with \newline persons\vspace{6pt}} ,Naive}]
            \addlegendimage{fill=laser_red, laser_red,sharp plot,area legend}
            \addlegendimage{fill=wc_c, wc_c,sharp plot,area legend}
            \addlegendimage{fill=wa_c, wa_c,sharp plot,area legend}
            \addlegendimage{fill=wa_c, white,sharp plot,area legend}
            \addlegendimage{fill=wa_c, white,sharp plot,area legend}
            \addlegendimage{fill=wa_c, white,sharp plot,area legend}
            \addlegendimage{sharp plot, ultra thick}
            \addlegendimage{mark=*, mark size=1pt, only marks}
            \addlegendimage{mark=x, mark size=2pt, only marks}
            \addlegendimage{dotted, sharp plot, ultra thick}
            \end{customlegend}
            \end{tikzpicture}
        \end{subfigure}%
        \caption{Performance comparison of DROW, the Gandalf detector~\cite{Weinrich14ROMAN}, and a naive deep learning baseline on two test sets (a) and (b). As explained in Section~\ref{sec:baselines}, the naive baseline is not applicable on the Reha test set.}
        \label{fig:drow_performance}
    \end{figure*}
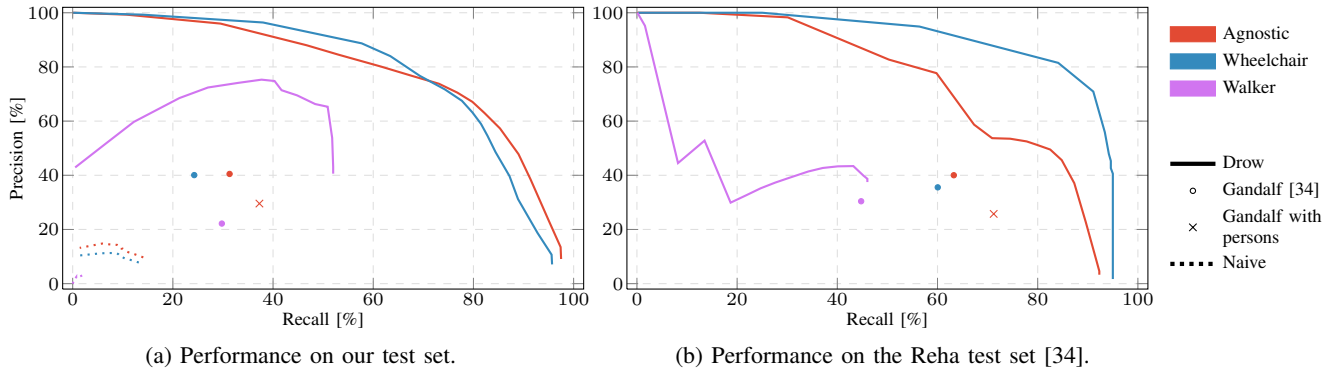

    We annotate our data with wheelchair and walker centers.
    Since the dataset contains 464k raw laser scans, we devise an annotation scheme that keeps effort manageable while still covering the full extent of the sequences as well as allowing temporal approaches to be developed in the future.
    Instead of all the laser scans, we annotate small batches throughout every sequence as follows:
    A batch consists of 100 frames, out of which we annotate every 5th frame, resulting in 20 annotated frames per batch.\footnote{This allows experimenting with interpolation between the annotations, even though we didn't do so in this work.}
    Within a sequence, we only annotate every 4th batch, leading to a total of \SI{5}{\percent} of the laser scans being annotated.
    We wrote a custom annotation tool based on matplotlib~\cite{Hunter07CSaE}, which loads sequences of scans with corresponding RGB images from the head camera and automatically finds the batches that need to be annotated.
    To aid the user in annotation, we show the first, current, and last image of the current batch.
    An example view of this annotation tool can be seen in Fig.\,\ref{fig:annotation_tools}.
    A \SI{1.2}{\m} circle around the mouse pointer, indicating the average wheelchair size, helps the user click on locations of wheelchair or walker centers.
    By using all of this supportive information to get an understanding of the scene, we were able to annotate most mobility aids, but based on the limited camera view, we likely still missed a few.

    \subsection{Training Procedure}
    We train a CNN which, for each preprocessed window, predicts whether it is indicative of a nearby wheelchair or walker and, if it is, predicts the offset of its center.
    We do so in a single pass by using a network with two output layers: a three-way SoftMax differentiating between \mytt{background}, \mytt{wheelchair} and \mytt{walker}, and a two-dimensional linear regression output.
    We optimize the network by minimizing the sum of a negative log-likelihood criterion on the SoftMax output and a root-mean-square error on the regression output.
    The regression targets are computed as $(\Delta x, \Delta y)$ in each window's local coordinate system as shown in Fig.\,\ref{fig:depth_prepocessing}.
    The class-labels are determined by the type of the closest detection to the window's center-point, with a maximal Euclidean distance of \SI{0.6}{\m} for wheelchairs and \SI{0.4}{\m} for walkers.
    When there is no nearby annotation, no error is backpropagated for the regression output and the network is thus free to predict any offset.

    The architecture of our network, inspired by the popular VGGnet~\cite{Simonyan15ICLR}\footnote{VGG is a well-performing architecture in vision. Other architectures may perform better, but that's beyond the scope of this paper.}, is as follows:
    Conv\,5@64, Conv\,5@64, Max\,2, Conv\,5@128, Conv\,3@128, Max\,2, Conv\,5@256, Conv\,3@5, where Conv\,$n$@$c$ represents a convolution with $c$ filters of size $n$ and Max\,$p$ the maximum operation on a window of $p$ values.
    Batch-normalization~\cite{Ioffe15Arxiv}, dropout~\cite{Srivastava14JMLR} of $0.25$, and ReLU nonlinearities~\cite{Glorot11AISTATS} are applied between all convolutional layers.
    For an input window resampled to $48$ values, the network outputs a vector of length five, three of which are sent to a SoftMax and the other two are the regression outputs $\Delta x$ and $\Delta y$.
    All weights are initialized similarly to the scheme proposed in~\cite{Saxe14ICLR}, except for those of the output layer, which are initialized to $0$.
    For training the network, we use the AdaDelta optimizer with $\rho=0.95$ and $\epsilon=10^{-7}$ and train until approximate convergence.\footnote{This means that we started training in the evening and stopped it the next morning. The loss-curve was almost flat in log-space.}
    During training, we add small multiplicative random noise to the regression targets and we flip each window and its target with probability $0.5$.
    We implemented our CNN in Theano~\cite{Bastien12NIPSW}.

    For the voting scheme, we add multiplicative class-weights to the votes and re-normalize them, thus voting for class $c$ with ${w_c p(c|w)} \over {\sum_{i \in \mathcal{C}} w_i p(i|w)}$ instead of $p(c|w)$.
    We then optimize all voting hyper-parameters (class-weights $\mathbf{w}$, grid resolution $b$ and blur-size $\sigma$) using hyperopt~\cite{Bergstra13ICML} to maximize $\max_T f_{\text{wheelchair}}(T) + f_{\text{walker}}(T)$ on our validation set, $f_{c}(T)$ being the F1-score of class $c$ using detection threshold $T$.
    Interestingly, the best values, $w_\text{bg}=0.38$, $w_\text{wheelchair}=0.60$, $w_\text{walker}=0.49$, $b=\SI{0.1}{\m}$ and $\sigma=2.93$ are not far off from our initial guesses.

    \subsection{Approach Evaluation}
    We trained our model on our training set and computed hyperparameters on our validation set as described in the previous section.
    In order to evaluate real-world performance of DROW, we now look at the precision and recall curves on our own test set and on the publicly available Reha test set of~\cite{Weinrich14ROMAN}\footnote{Detection annotations for the Reha dataset were provided by the authors. We skip the Home test set as it does not contain mobility aids.} recorded with a similar robot.
    Recall that our test set was recorded in a never before seen part of the care facility, thus a method which learns location priors or background models will fail.
    The result shown in Fig.\,\ref{fig:drow_performance} demonstrates that DROW generalizes very well and has significantly higher precision and recall than Gandalf~\cite{Weinrich14ROMAN}, both on our test set and on the Reha test set.

    In our application scenario, the actual detection of a mobility aid is more important than their correct classification, hence we also evaluate a class-agnostic performance.
    For this, we ignore the class of all detections and annotations during evaluation.

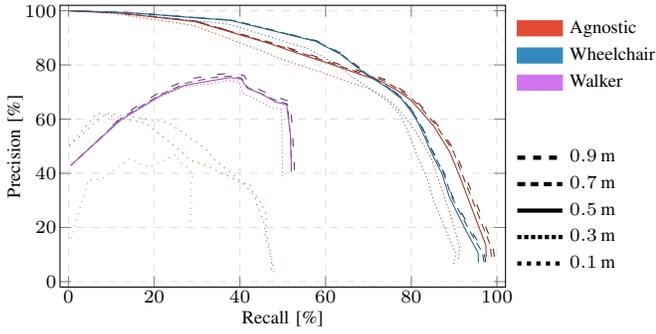
\begin{figure}
        \begin{subfigure}{0.79\linewidth}
            \begin{tikzpicture}
            \begin{axis}[precrec, height=5.33cm]
            %Nice overview of dashes, could also do custom ones here! http://tex.stackexchange.com/questions/45275/tikz-get-values-for-predefined-dash-patterns
            \addplot [loosely dashed, mark=none,  color=laser_red!44!black] table [y=p_all_0.9, x=r_all_0.9, col sep=comma] {data/pr_test.csv};\label{r0.9}
            \addplot [loosely dashed, mark=none,  color=wc_c!44!black] table [y=p_wc_0.9, x=r_wc_0.9, col sep=comma] {data/pr_test.csv};
            \addplot [loosely dashed, mark=none,  color=wa_c!44!black] table [y=p_wa_0.9, x=r_wa_0.9, col sep=comma] {data/pr_test.csv};
            \addplot [densely dashed, mark=none,  color=laser_red!58!black] table [y=p_all_0.7, x=r_all_0.7, col sep=comma] {data/pr_test.csv};\label{r0.7}
            \addplot [densely dashed, mark=none,  color=wc_c!58!black] table [y=p_wc_0.7, x=r_wc_0.7, col sep=comma] {data/pr_test.csv};
            \addplot [densely dashed, mark=none,  color=wa_c!58!black] table [y=p_wa_0.7, x=r_wa_0.7, col sep=comma] {data/pr_test.csv};
            \addplot [solid, mark=none, color=laser_red!72!black] table [y=p_all_0.5, x=r_all_0.5, col sep=comma] {data/pr_test.csv};\label{r0.5}
            \addplot [solid, mark=none, color=wc_c!72!black] table [y=p_wc_0.5, x=r_wc_0.5, col sep=comma] {data/pr_test.csv};
            \addplot [solid, mark=none, color=wa_c!72!black] table [y=p_wa_0.5, x=r_wa_0.5, col sep=comma] {data/pr_test.csv};
            \addplot [densely dotted, mark=none, color=laser_red!86!black] table [y=p_all_0.3, x=r_all_0.3, col sep=comma] {data/pr_test.csv};\label{r0.3}
            \addplot [densely dotted, mark=none, color=wc_c!86!black] table [y=p_wc_0.3, x=r_wc_0.3, col sep=comma] {data/pr_test.csv};
            \addplot [densely dotted, mark=none, color=wa_c!86!black] table [y=p_wa_0.3, x=r_wa_0.3, col sep=comma] {data/pr_test.csv};
            \addplot [dotted, mark=none, color=laser_red] table [y=p_all_0.1, x=r_all_0.1, col sep=comma] {data/pr_test.csv};\label{r0.1}
            \addplot [dotted, mark=none, color=wc_c] table [y=p_wc_0.1, x=r_wc_0.1, col sep=comma] {data/pr_test.csv};
            \addplot [dotted, mark=none, color=wa_c] table [y=p_wa_0.1, x=r_wa_0.1, col sep=comma] {data/pr_test.csv};
            \end{axis}
            \end{tikzpicture}
        \end{subfigure}\hspace*{-5pt}%
        \begin{subfigure}{0.2\linewidth}
            \vspace*{-15pt}
            \begin{tikzpicture}
            \begin{customlegend}[every axis legend/.append style={nodes={right}}, legend style={draw=none,font=\scriptsize}, legend entries={Agnostic, Wheelchair, Walker,~,~,~,\SI{0.9}{\m}, \SI{0.7}{\m}, \SI{0.5}{\m}, \SI{0.3}{\m}, \SI{0.1}{\m}}]
            \addlegendimage{fill=laser_red, laser_red,sharp plot,area legend}
            \addlegendimage{fill=wc_c, wc_c,sharp plot,area legend}
            \addlegendimage{fill=wa_c, wa_c,sharp plot,area legend}
            \addlegendimage{fill=wa_c, white,sharp plot,area legend}
            \addlegendimage{fill=wa_c, white,sharp plot,area legend}
            \addlegendimage{fill=wa_c, white,sharp plot,area legend}
            \addlegendimage{dashed, sharp plot, ultra thick}
            \addlegendimage{densely dashed, sharp plot, ultra thick}
            \addlegendimage{solid, sharp plot, ultra thick}
            \addlegendimage{densely dotted, sharp plot, ultra thick}
            \addlegendimage{dotted, sharp plot, ultra thick}
            \end{customlegend}
            \end{tikzpicture}
        \end{subfigure}%
        \caption{Performance of DROW for different evaluation radii. The high overlap for radii above 0.3 shows that our predictions are well localized.}
        \label{fig:radii}
    \end{figure}

    The precision and recall curves for each class are computed by varying the voting threshold $T$.
    We use an evaluation radius of \SI{0.5}{\m}, meaning a detection is matched to a ground-truth if it falls within \SI{0.5}{\m} of that ground-truth's center and has the correct class.
    An annotation can be matched with at most one detection, all remaining detections of that class are false positives and all unmatched annotations of that class are false negatives.
    To see how well localized DROW's detections are, we show precision-recall curves for varying evaluation radii in Fig.\,\ref{fig:radii}.
    The cluttering of all curves for acceptance radii of \SI{0.3}{\m} and above mean that our detections are well localized.
    During annotation, it became clear to us that the precision of the annotations is limited, meaning that evaluation at radii as small as \SI{0.1}{\m} is strongly affected by labeling noise.

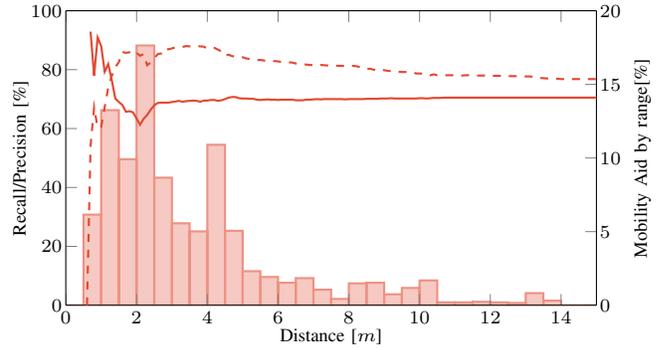
\begin{figure}[t]
        \def\selectedthresh{0.500000}
        \begin{tikzpicture}
        \begin{axis}[precrecbase,
        yticklabel style = {xshift=0.6ex},
        xmin = 0, xmax = 15,
        ymin = 0, ymax = 20.,
        hide x axis,
        axis y line*=right,
        ylabel near ticks,
        ylabel={Mobility Aid by range[\%]},
        ylabel style={font=\scriptsize,yshift=-5pt},
        height=5.5cm,
        width=\linewidth,
        ybar, ybar interval,
        legend image post style={line width =0.3pt}]
        \addplot [laser_red, opacity=0.6, thick, fill=laser_red, fill opacity=0.3] table [y=percentage, x=dist, col sep=comma] {data/test_density.csv};\label{test_density}
        \end{axis}
        \begin{axis}[precrecbase,
        xticklabel style = {font=\scriptsize,yshift=-0.2ex},
        xmin = -0, xmax = 15,
        ymin = 0, ymax = 100,
        axis y line*=left,
        height=5.5cm,
        width=\linewidth,
        xlabel={Distance [$m$]},
        ylabel={Recall/Precision [\%]}]
        \addplot [laser_red, thick,] table [y=p_all, x=d, col sep=comma] {data/pr_dist_test-\selectedthresh.csv};\label{test_p_a}
        \addplot [dashed, laser_red, thick,] table [y=r_all, x=d, col sep=comma] {data/pr_dist_test-\selectedthresh.csv};\label{test_r_a}
        %\addplot [wc_c, thick,] table [y=p_wc, x=d, col sep=comma] {data/pr_dist_test-\eer.csv};\label{test_p_wc}
        %\addplot [dashed, wc_c, thick,] table [y=r_wc, x=d, col sep=comma] {data/pr_dist_test-\eer.csv};\label{test_r_wc}
        %\addplot [wa_c, thick,] table [y=p_wa, x=d, col sep=comma] {data/pr_dist_test-\eer.csv};\label{test_p_wa}
        %\addplot [dashed, wa_c, thick,] table [y=r_wa, x=d, col sep=comma] {data/pr_dist_test-\eer.csv};\label{test_r_wa}
        \end{axis}
        \end{tikzpicture}
        \caption{Precision (\rf{test_p_a}) and Recall (\rf{test_r_a}) at $T=0.5$ in the class-agnostic case up to a certain distance in meters. The histogram (\,\protect\raisebox{-2pt}{\rf{test_density}}\,, y axis on the right) shows the percentage of all annotations at a certain range.}
        \label{fig:distance_eval}
    \end{figure}

    We also analyze how well DROW behaves with respect to the distance from the laser scanner.
    For this, we start by ignoring all detections beyond \SI{0.1}{\m} from the laser.
    We then slowly grow that radius, taking into account more and more detections, and plot how the precision and recall at a fixed threshold evolve in Fig.\,\ref{fig:distance_eval}.
    DROW performs very well across the board, but especially so in mid-range which is crucial for navigation and planning.
    Unsurprisingly, the curves do not change much beyond a distance of \SI{10}{\m}, as only few mobility aids where observed that far.

    \subsection{Baselines}\label{sec:baselines}
    We compare our proposed method to two baselines in Fig.\,\ref{fig:drow_performance}: the publicly available Gandalf detector~\cite{Weinrich14ROMAN} and a naive deep learning baseline.

    None of the precision-recall curves in~\cite{Weinrich14ROMAN} quantify Gandalf's actual detection performance, they only focus on parts of the system independently.
    To obtain comparable detection precision-recall values, we evaluate Gandalf\footnote{\url{https://github.com/neurob/gandalf_detector}} using the evaluation protocol described above.
    As provided, the Gandalf detector has no single threshold to be tuned and thus results in a single point in our plots.
    For the class-agnostic case, we plot the results for the cases when Gandalf person detections are kept~(\,\rf{plot:gandalf}\,) and discarded~(\,\rf{plot:gandalfp}\,), respectively, showing a trade-off between precision and recall.

    Note that the detector performance we obtained with the code from~\cite{Weinrich14ROMAN} is significantly lower than the one reported in the original paper.
    The performance could be improved by an additional covariance-based merging step briefly mentioned in~\cite{Weinrich14ROMAN} which is, however, neither described in detail in the paper, nor included in the provided detector code.
    We were thus not able to reproduce the results presented in~\cite{Weinrich14ROMAN}.
    However, extrapolating from the false positives and missing detections we observed using the provided code, even an optimistic bound on this method's performance would place it below our detector's curve.

    As a naive deep learning baseline, we train another CNN, similarly to YOLO~\cite{Redmon15Arxiv}, that directly predicts up to two detections (sufficient for \SI{95.0}{\percent} of the frames) in normalized $(x,y)$ coordinates, based on a full scan.
    The results of this baseline experiment are shown as dotted lines in Fig.\,\ref{fig:drow_performance}~(a), but are missing from the Reha test set as the scans have a different amount of laser beams.
    It should be noted that we spent a considerable amount of time getting the naive CNN baseline to work as well as possible and follow best-practice: carefully chosen receptive-fields, batch-normalization, careful initialization, AdaDelta optimizer,~\etc.
    However, as can easily be seen, just applying deep learning to 2D range data in such a naive way leads to abysmal results.

    \subsection{Ablation Studies}
    \label{sec:ablation}
    In this section, we analyze how each of our design decisions affects detection by systematically removing or substituting each part that defines our approach.
    Fig.\,\ref{fig:abblation} summarizes these experiments.
    Overall, the complete DROW detector (\rf{plot:us}) outperforms all other configurations, \ie each single decision contributes to the high performance.
    For each of the following experiments, we retrain the full network from scratch.

    \newcommand{\addsetup}[4]{
        \addplot [thick,solid,  mark=none, color=#2] table [y=p_#4_0.5, x=r_#4_0.5, col sep=comma] {data/#1.csv};\label{#3}
    }
    \newcommand{\addall}[1]{
        \addsetup{pr_test}{good_gray}{plot:us}{#1}
        \addsetup{pr_test_nocent}{awesome_orange}{plot:no_centering}{#1}
        \addsetup{pr_test_noclamp}{wc_c}{plot:no_clamping}{#1}
        \addsetup{pr_test_raw-vote}{wa_c}{plot:no_preproc}{#1}
        \addsetup{pr_test_fixangle}{bg_green}{plot:no_resizing}{#1}
        \addsetup{pr_test_absR}{laser_red}{plot:no_delta_r}{#1}
        \addsetup{pr_test_wcnet-mlp}{shit_brown}{plot:mlp}{#1}
        \addsetup{pr_test_nothing}{deep_pink}{plot:nothing}{#1}
        \addsetup{pr_test_rf_baseline}{random_blue}{plot:rf}{#1}
    }

    \begin{figure*}
        \begin{subfigure}{0.33\textwidth}
        \begin{tikzpicture}
            \begin{axis}[precrec, height=5cm, width=\linewidth]
            \addall{all}
            \end{axis}
        \end{tikzpicture}
        \caption{Class-agnostic}
        \end{subfigure}\hspace*{-15pt}%
        \begin{subfigure}{0.33\textwidth}
        \begin{tikzpicture}
            \begin{axis}[precrec, height=5cm, width=\linewidth, ylabel={}]
            \addall{wc}
            \end{axis}
        \end{tikzpicture}
        \caption{Wheelchairs}
        \end{subfigure}\hspace*{-23pt}%
        \begin{subfigure}{0.33\textwidth}
        \begin{tikzpicture}
            \begin{axis}[precrec, height=5cm, width=\linewidth, ylabel={}]
            \addall{wa}
            \end{axis}
        \end{tikzpicture}
        \caption{Walkers}
        \end{subfigure}\hspace*{-22pt}%
        \begin{subfigure}{0.10\textwidth}
        \vspace*{-30pt}
        \begin{tikzpicture}
        \begin{customlegend}[every axis legend/.append style={nodes={right}}, legend style={draw=none,font=\scriptsize}, legend entries={DROW, \tiny~,  No centering, No clamping, No resampling, Raw window,\tiny~, Predict {$r, \Delta \varphi$}, Basic,\tiny~, MLP, Random Forest}]
            \addlegendimage{good_gray,sharp plot, ultra thick}
            \addlegendimage{fill=wa_c, white,sharp plot}
            \addlegendimage{awesome_orange,sharp plot, ultra thick}
            \addlegendimage{wc_c,sharp plot, ultra thick}
            \addlegendimage{bg_green,sharp plot, ultra thick}
            \addlegendimage{wa_c,sharp plot, ultra thick}
            \addlegendimage{fill=wa_c, white,sharp plot}
            \addlegendimage{laser_red,sharp plot, ultra thick}
            \addlegendimage{deep_pink,sharp plot, ultra thick}
            \addlegendimage{fill=wa_c, white,sharp plot}
            \addlegendimage{shit_brown,sharp plot, ultra thick}
            \addlegendimage{random_blue,sharp plot, ultra thick}
        \end{customlegend}
        \end{tikzpicture}
        \end{subfigure}%
        \caption{Ablation studies. We retrain our approach with slight modifications to show how the performance changes.}
        \label{fig:abblation}
    \end{figure*}
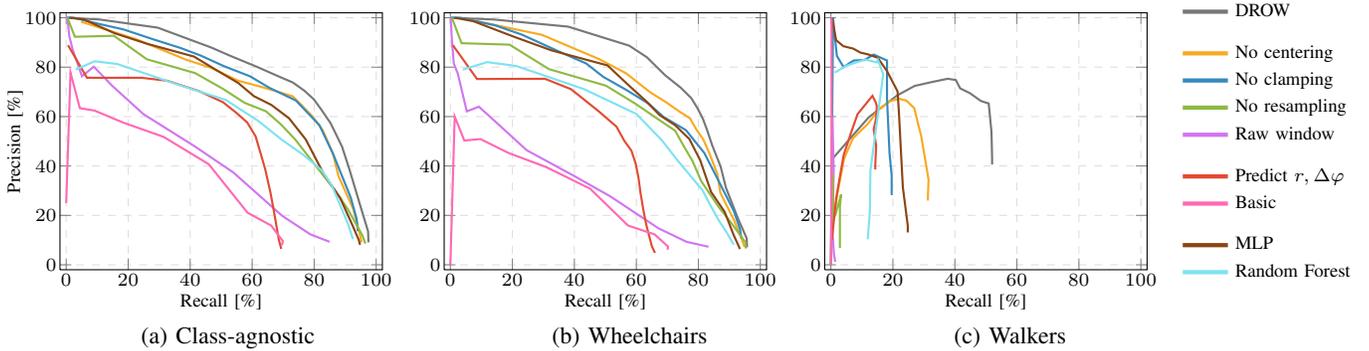

    \subsubsection{Preprocessing}
    Our depth preprocessing can be dissected into three operations: centering, clamping and resampling.
    Not centering the input window (\rf{plot:no_centering}) has the smallest, albeit non-negligible, effect.
    Removing the clamping (\rf{plot:no_clamping}) hurts overall performance as much as removing the centering.
    Without the resampling step (\rf{plot:no_resizing}) which ensures that the network's receptive field keeps a constant real-world size, performance drops significantly, especially for walkers.
    Considering that the CNN has to learn completely different representations across different distances, the overall performance is still surprisingly good.
    The fact that walkers are almost undetected and yet the agnostic performance is decent suggests a high confusion.
    Finally, dropping all of the above steps (\rf{plot:no_preproc}) performs the worst in all measures, showing that our preprocessing is indeed crucial.

    In total, all these experiments clearly suggest that each of the three preprocessing steps are important.

    \subsubsection{Voting}
    The comparison to the naive deep learning baseline in Fig.\,\ref{fig:drow_our_test} already showed that voting is essential for proper detection results.

    In order to see the effect of regressing local offsets $(\Delta x, \Delta y)$, we train a version of the network which regresses $(\Delta\varphi, r)$: an angular offset and an absolute distance to the laser scanner (\rf{plot:no_delta_r}).
    For this experiment, we also had to remove centering (\cf~\rf{plot:no_centering}), as otherwise $r$ is impossible to predict.
    Without centering, the network could in principle learn to ``pass through'' the centerpoint's distance and add it as a bias in the last layer.
    The drastically degraded results suggest that this is a much more difficult problem that the learning procedure fails to solve.

    To ensure that this difficulty is not caused by a bad interaction with our preprocessing, we also train a network to make these predictions on raw, unprocessed data (\rf{plot:nothing}).
    This model performing worst of all shows that even a plain voting-based network alone does not perform well if no care is taken on both the input and output.

    \begin{figure}[b]
        \begin{subfigure}{0.77\linewidth}
            \begin{tikzpicture}
            \begin{axis}[precrec, height=5.33cm]
            %Nice overview of dashes, could also do custom ones here! http://tex.stackexchange.com/questions/45275/tikz-get-values-for-predefined-dash-patterns
            \addplot [loosely dashed, mark=none,  color=laser_red, thick] table [y=p_all_0.5, x=r_all_0.5, col sep=comma] {data/pr_reha_2.2m-64.csv};
            \addplot [loosely dashed, mark=none,  color=wc_c, thick] table [y=p_wc_0.5, x=r_wc_0.5, col sep=comma] {data/pr_reha_2.2m-64.csv};
            \addplot [loosely dashed, mark=none,  color=wa_c, thick] table [y=p_wa_0.5, x=r_wa_0.5, col sep=comma] {data/pr_reha_2.2m-64.csv};
            \addplot [solid, mark=none, color=laser_red, thick] table [y=p_all_0.5, x=r_all_0.5, col sep=comma] {data/pr_test.csv};
            \addplot [solid, mark=none, color=wc_c, thick] table [y=p_wc_0.5, x=r_wc_0.5, col sep=comma] {data/pr_test.csv};
            \addplot [solid, mark=none, color=wa_c, thick] table [y=p_wa_0.5, x=r_wa_0.5, col sep=comma] {data/pr_test.csv};
            \addplot [dotted, mark=none, color=laser_red, thick] table [y=p_all_0.5, x=r_all_0.5, col sep=comma] {data/pr_reha_1.1m-32.csv};
            \addplot [dotted, mark=none, color=wc_c, thick] table [y=p_wc_0.5, x=r_wc_0.5, col sep=comma] {data/pr_reha_1.1m-32.csv};
            \addplot [dotted, mark=none, color=wa_c, thick] table [y=p_wa_0.5, x=r_wa_0.5, col sep=comma] {data/pr_reha_1.1m-32.csv};
            \end{axis}
            \end{tikzpicture}
        \end{subfigure}\hspace*{-5pt}%
        \begin{subfigure}{0.2\linewidth}
            \vspace*{-15pt}
            \begin{tikzpicture}
            \begin{customlegend}[every axis legend/.append style={nodes={right}}, legend style={draw=none,font=\scriptsize}, legend entries={Agnostic, Wheelchair, Walker,~,~,~,$\ell=\SI{2.2}{\m}$, $\ell=\SI{1.66}{\m}$, $\ell=\SI{1.1}{\m}$}]
            \addlegendimage{fill=laser_red, laser_red,sharp plot,area legend}
            \addlegendimage{fill=wc_c, wc_c,sharp plot,area legend}
            \addlegendimage{fill=wa_c, wa_c,sharp plot,area legend}
            \addlegendimage{fill=wa_c, white,sharp plot,area legend}
            \addlegendimage{fill=wa_c, white,sharp plot,area legend}
            \addlegendimage{fill=wa_c, white,sharp plot,area legend}
            \addlegendimage{dashed, sharp plot, ultra thick}
            \addlegendimage{solid, sharp plot, ultra thick}
            \addlegendimage{dotted, sharp plot, ultra thick}
            \end{customlegend}
            \end{tikzpicture}
        \end{subfigure}%
        \caption{Effect of the window size $\ell$ on the performance.}
        \label{fig:window_size}
    \end{figure}
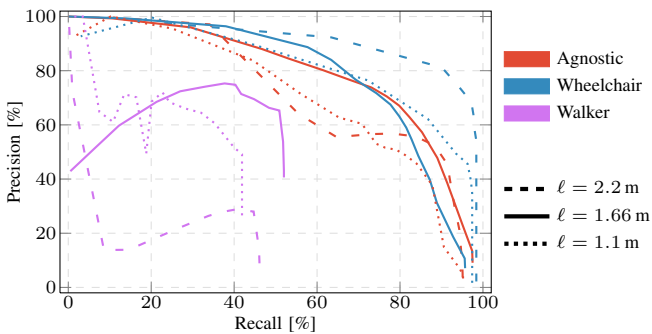

    \subsubsection{Model}
    Since every window is sent through the model individually, and all windows have the same size due to the resampling step, there is no inherent reason for the model to be a CNN.
    To show that the spatial prior encoded in a CNN is useful, we train a three hidden-layer perceptron (\rf{plot:mlp}) with $2048$ hidden units each, ReLU non-linearities, dropout, and batch-normalization.
    As in the other experiments, we leave all other design decisions unchanged.
    The result shows that, although the MLP is a viable alternative, the CNN clearly outperforms it.

    As an additional baseline we trained a regression forest (\rf{plot:rf}) that regresses the three class probabilities and the local offsets $(\Delta x, \Delta y)$.
    It was trained with the same training data as the CNN and the MLP (including the flip augmentation).
    We used the scikit-learn~\cite{Pedregosa11JMLR} implementation of the regression forest with all default settings and 50 trees.
    Training took 7.5 hours (using 5 threads) and resulted in a model of 6.9GB compared to our CNN model of 1.1MB.
    Nevertheless, the forest model performs even slightly worse than the MLP in all cases.

    \subsubsection{Window Size}
    So far, all evaluations have been performed with a real-world window spanning $\ell=\SI{1.66}{\m}$, motivated by the typical extent of wheelchairs being \SI{1.2}{\m}.
    To verify that this is a reasonable choice, we trained a network on both larger (\SI{2.20}{\m}) and smaller (\SI{1.10}{\m}) windows.
    The results of this experiment can be seen in Fig.\,\ref{fig:window_size}.
    The class-agnostic prediction performance is best for our choice of $\ell=\SI{1.66}{\m}$, but wheelchairs alone could be detected considerably better by using a window of \SI{2.2}{\m}.
    These results suggest that further improvements might be obtained by providing the network with multiple scales of input windows.

    %%%%%%%%%%%%%%%%%%%%%%%%%%%%%%%%%%%%%%%%%%%%%%%%%%%%%%%%%%%%%%%%%%%%%%%%%%%%%%%%
    \section{RELATED WORK}
    Most related to our approach is that of Weinrich~\etal~\cite{Weinrich14ROMAN} who propose a distance invariant feature for laser based detection.
    They detect wheelchairs, walkers, and persons in 2D range data by first detecting larger jumps in the distance.
    Once such a jump is found, they create a window with a fixed real-world extent, covering subsequent scan points.
    The window is discretized into equally sized segments, for each of which a clamped distance relative to the window depth is computed.
    Each segment is characterized by the min, max and average depth which are concatenated for all segments to form the feature vector.
    These are then classified by an AdaBoost classifier.
    While their feature design has a high similarity to our depth preprocessing step, their windows are only created when a jump in distance is found, where we create a window for every scan point.
    The biggest difference, however, is that they directly predict a detection center and object class for every window, while we vote for centers, a step in our detector which we have shown to be essential.

    A few more publications~\cite{DeDeuge13ACRA,Lai14ICRA,Maturana15IROS} identify the need for a distance-invariant representation which we highlight in this work.
    The solution of creating a re-sampled depth image from a 3D point-cloud by ray-tracing~\cite{DeDeuge13ACRA}, however, involves multiple significantly more complex operations than our proposed simple pre-processing and no timings were provided.
    The other commonly-used solution is the creation of a 3D occupancy-grid~\cite{Lai14ICRA,Maturana15IROS}.
    Such an approach effectively uses $N+1$-dimensional inputs for data on an $N$-dimensional manifold, while our approach keeps the input $N$-dimensional.
    Arguably, the latter makes learning easier and predictions more robust~\cite{Bengio13TPAMI}.
    In addition, multiple different heuristics exist for ``filling holes'' in those occupancy-grids~\cite{Maturana15IROS}, whereas our pre-processing doesn't produce holes in the first place.
    In the end, both data representations may be made to work, but we believe our lower-dimensional representation is simpler and more effective.

    Others have used voting for detection in laser scans, Mozos~\etal~\cite{Mozos10IJSR} in a multi-height laser setup and Spinello~\etal~\cite{Spinello10AAAI} in a layered fashion based on 3D range data.
    Both, related to~\cite{Leibe08IJCV}, learn shape models from data and cast votes from the different laser layers to detection centroids.
    However, similar to Weinrich~\etal~\cite{Weinrich14ROMAN}, they rely on jump distance segmentation and only the segments can cast votes for detections.
    In~\cite{Wang15RSS}, Wang~\etal~{\em prove\/} that detection using a voting scheme corresponds to sliding-window detection on a sparse grid {\em for linear models}.
    While they achieve good detections on point clouds using hand-crafted features and a linear SVM, it has been shown time and again in the computer vision literature (and in this paper) that learned non-linear features and classifiers outperform hand-crafted features and linear classifiers by a large margin.
    In line with our findings, voting has recently been shown to work well in combination with various other deep learning approaches~\cite{Milletari16Arxiv,Riegler14BMVC,Xie15MICCAI}.
    It would thus be interesting to establish a relationship between deep, non-linear voting detectors such as DROW and sliding-window detectors on sparse inputs as shown in~\cite{Wang15RSS}.

    Detection in range data is not limited to persons or mobility aids.
    Around the time at which we uploaded the preprint\footnote{\url{https://arxiv.org/abs/1603.02636v1}} of this paper, Ondruska~\etal~\cite{Ondruska16RSSWorkshop} shows how to do ``tracking'' of pedestrians, cyclists, buses, cars, and road obstacles in 2D range data.
    Although seemingly similar to our work, their input is an occupancy-grid (i.e. $N+1$-dimensional) for which they predict a labelled occupancy-grid, as opposed to discrete detections or tracks with IDs.
    Based on the biases for static grid cells in their RNN, it remains to be seen how well their approach works on a mobile robot.

    Merdrignac~\etal~\cite{Merdrignac15ARSO} also detect cars, bicyclists and static road obstacles in 2D range data.
    They hypothesize that more information can be extracted from range data than done before and aim to achieve this by designing a large set of hand-crafted features.
    We agree, but we instead learn feature representations from the data directly.

    %%%%%%%%%%%%%%%%%%%%%%%%%%%%%%%%%%%%%%%%%%%%%%%%%%%%%%%%%%%%%%%%%%%%%%%%%%%%%%%%
    \section{DISCUSSION}
    A few interesting aspects of our detector should be highlighted.
    Firstly, given the experimental evaluation it becomes clear that we perform well with respect to wheelchairs, but our walker performance leaves some room for improvement.
    One reason for this could be the fact that our training set is rather biased towards wheelchairs.
    However, the class-agnostic precision-recall curves are very similar to the ones of wheelchairs in most of our experiments.
    This suggests that one significant problem for walker detection is a high confusion with wheelchairs.
    The fact that whenever DROW detects a mobility aid of either class, it is well localized, leads us to the conclusion that the network ``understood'' walkers but needs to see more of them.

    Secondly, laser-based detection is often dismissed as too sparse or too difficult in a single frame.
    We hope our results refute this fear, even though we do not dismiss the performance gain that could likely be achieved by tracking our detections over time.

    A commonly suggested alternative to laser-based detections is the use of a vision-based detector and RGB(-D) cameras, which produce much richer data.
    Several approaches have tried this before~\cite{Myles02ACCV,Huang10TITB,Cauchois05IROS}, but none of them is general enough to be applied in all scenarios.
    They typically rely on geometric primitives~\cite{Myles02ACCV} or very specific camera setups~\cite{Cauchois05IROS}.
    Furthermore, commonly used RGB-D sensors such as the Asus Xtion, mounted on a human sized robot, have a too narrow field of view to perceive wheelchairs both far away, as well as when they come close to the robot.
    Vision-based detectors would thus need many cameras to cover the same field of view that is covered by a single laser scanner.
    Additionally, since mobility aids themselves come in many different appearances, training such a detector robustly would require a vastly larger dataset.
    Thus, a laser-based detector, as we have shown to be feasible here, is the more effective and efficient approach.

    %%%%%%%%%%%%%%%%%%%%%%%%%%%%%%%%%%%%%%%%%%%%%%%%%%%%%%%%%%%%%%%%%%%%%%%%%%%%%%%%
    \section{CONCLUSIONS}
    In this paper we introduced the DROW detector, a fast deep learning based detector for wheelchairs and walkers from 2D range data.
    We have proposed a depth preprocessing and a voting scheme, both of which enable CNNs to vastly outperform naive CNN detection baselines and obtain state of the art results compared to a previous method.
    We performed a thorough experimental evaluation, justifying all our major design choices.
    To achieve all this, we recorded a large dataset in which we annotated wheelchair and walker centroids, and which we believe will enable further research.
    We are convinced that our detector generalizes to other classes given training data and will thus be useful to the community.
    Upon acceptance of the paper, we will publish our code, including a ROS node and the annotated dataset.

    %\addtolength{\textheight}{-12cm}

    %%%%%%%%%%%%%%%%%%%%%%%%%%%%%%%%%%%%%%%%%%%%%%%%%%%%%%%%%%%%%%%%%%%%%%%%%%%%%%%%
    %\section*{ACKNOWLEDGMENT}
    %
    %Leave this for the final version
    %
    %%%%%%%%%%%%%%%%%%%%%%%%%%%%%%%%%%%%%%%%%%%%%%%%%%%%%%%%%%%%%%%%%%%%%%%%%%%%%%%%
    \bibliographystyle{plain}
    \bibliography{abbrev_short,egbib}

    \begin{figure}[b!]
    \centering
    \includegraphics[width=0.95\linewidth]{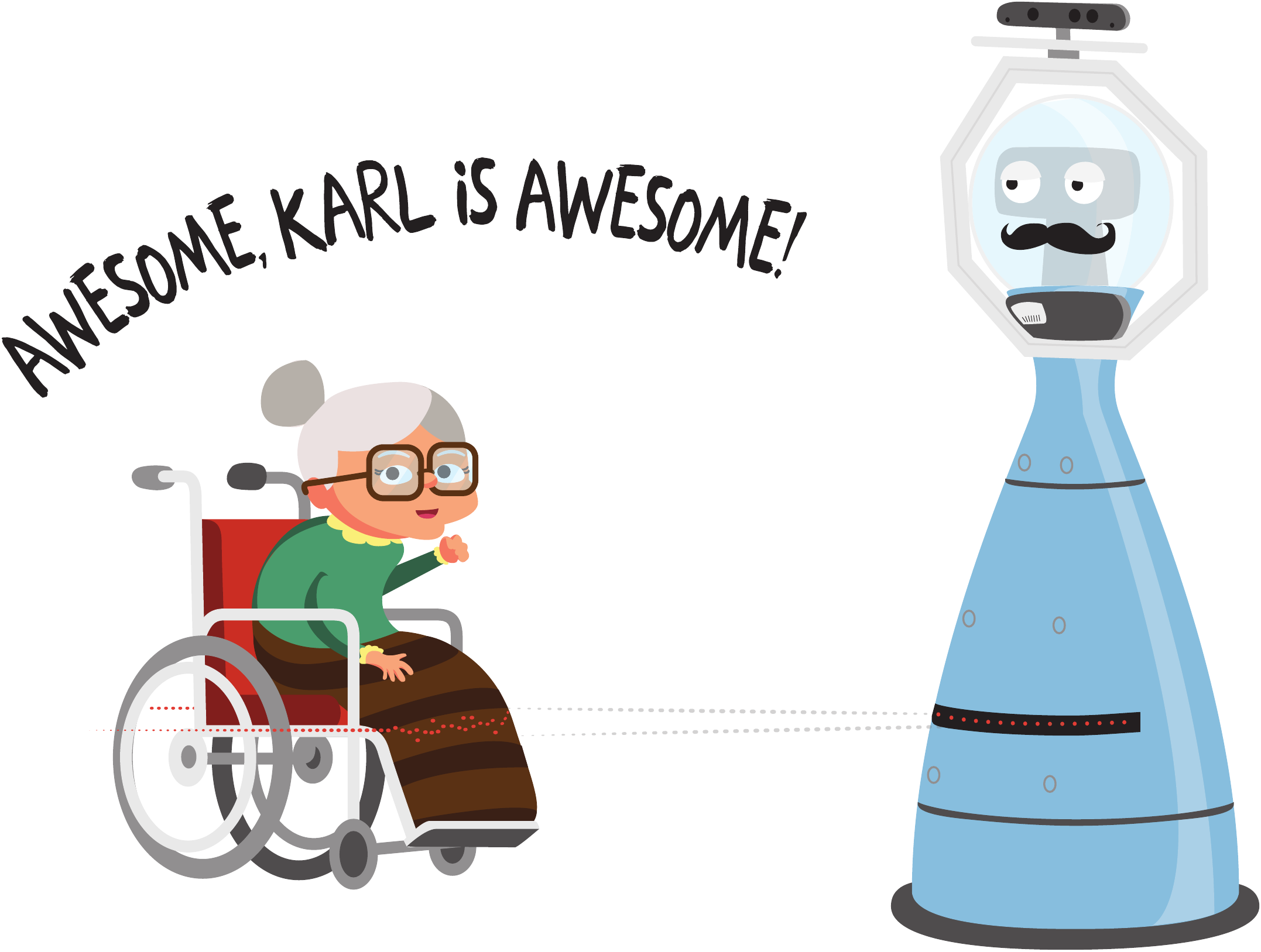}
    \caption*{Image credit: Thibault Jan Beyer}
    \end{figure}

\end{document}